\documentclass[sigconf,nonacm]{acmart}

\setcopyright{none}

\usepackage{booktabs}
\usepackage{multirow}
\usepackage{graphicx}
\usepackage{subcaption}
\usepackage{xcolor}
\usepackage{balance}
\usepackage{stfloats}
\usepackage{fancyhdr}
\usepackage{listings}

\usepackage{booktabs}
\usepackage{tabularx}
\usepackage{array}
\title{Recursive Agent Harnesses}

\author{Elias Lumer}
\affiliation{\institution{PricewaterhouseCoopers, U.S.}\country{}}
\email{elias.lumer@pwc.com}

\author{Sahil Sen}
\affiliation{\institution{PricewaterhouseCoopers, U.S.}\country{}}
\email{sahil.s.sen@pwc.com}

\author{Kevin Paul}
\affiliation{\institution{PricewaterhouseCoopers, U.S.}\country{}}

\author{Vamse Kumar Subbiah}
\affiliation{\institution{PricewaterhouseCoopers, U.S.}\country{}}

\renewcommand{\shortauthors}{Lumer et al.}

\begin{document}


\begin{abstract}
Recursive language models (RLMs) showed that recursion over model calls is an
effective strategy for long-context reasoning, and production coding agents have
begun to write code that spawns subagents at scale, most recently in Anthropic's
dynamic workflows. We name and study the pattern between these two lines of work,
where the recursive unit is a full agent harness with filesystem tools, code
execution, and planning rather than a model call with no tools. We call this the
\textbf{Recursive Agent Harness (RAH)} and frame it as \emph{harness recursion}, the
code-first extension to the \emph{model recursion} of RLMs. A parent agent generates
and runs an executable script that spawns subagent harnesses in parallel for
fine-grained workloads and uses structured function calls for small subtasks. We
provide a controlled evaluation on long-context reasoning. With the backbone held
fixed at GPT-5 to match the published Codex and RLM baselines, RAH improves the Codex
coding-agent baseline from 71.75\% to 81.36\% on Oolong-Synthetic (199 samples, 13
context-length buckets up to 4M tokens), a gain attributable to the harness rather
than the model. With a stronger backbone, Claude Sonnet 4.5, the same design reaches
89.77\%.
\end{abstract}


\maketitle

\fancypagestyle{plain}{
  \fancyhf{}
  \fancyfoot[C]{\thepage}
  \renewcommand{\headrulewidth}{0pt}
  \renewcommand{\footrulewidth}{0pt}
}

\fancypagestyle{senstyle}{
  \fancyhf{}
  \fancyhead[R]{Lumer et al.}
  \fancyfoot[C]{\thepage}
  \renewcommand{\headrulewidth}{0pt}
  \renewcommand{\footrulewidth}{0pt}
}

\thispagestyle{plain}
\pagestyle{senstyle}

\begin{figure*}[t]
  \centering
  \includegraphics[width=14cm]{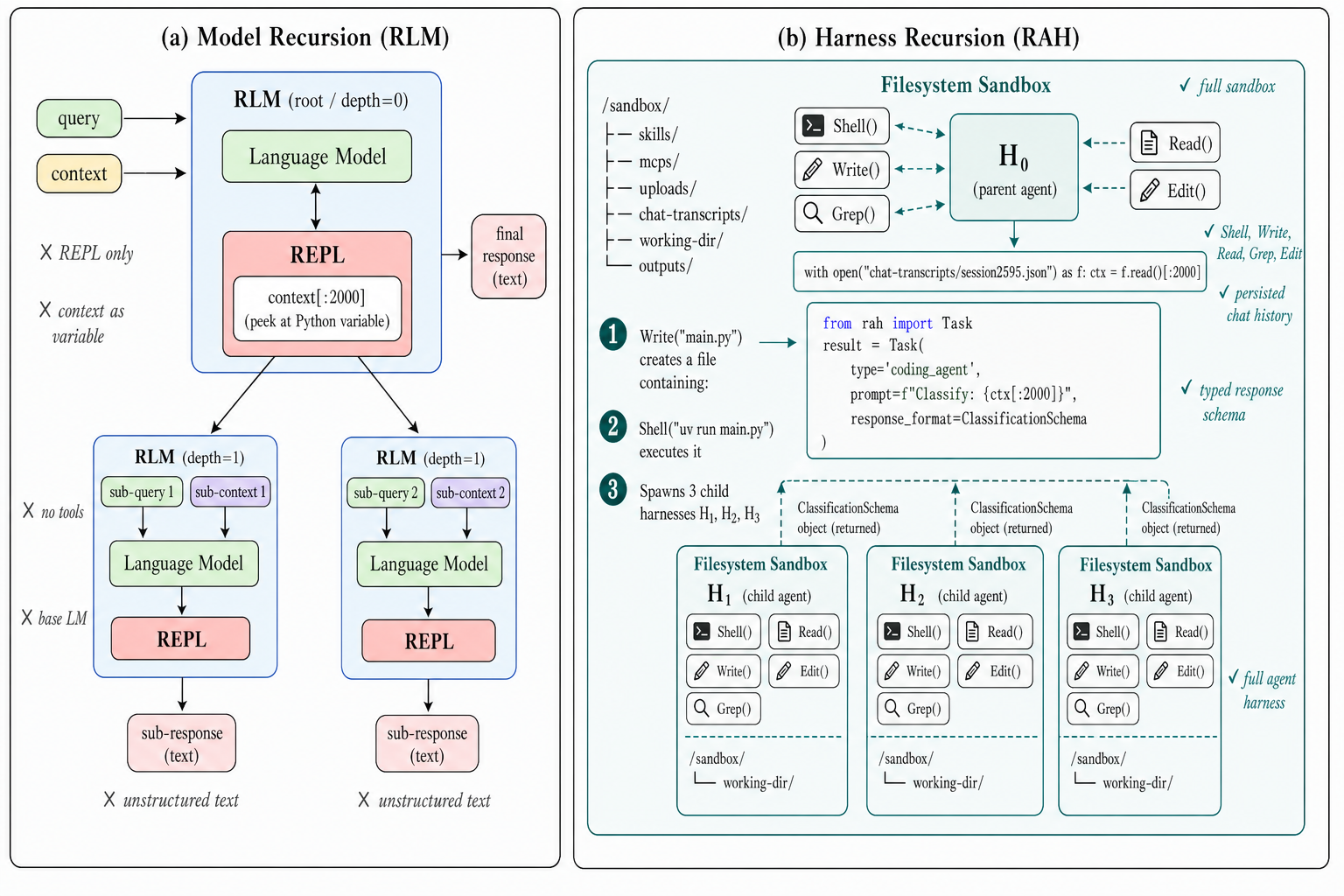}
  \caption{The Recursive Agent Harness (RAH). A parent agent selects between
  code-execution spawning (writing an executable script that spawns
  subagents in parallel) and JSON tool-call spawning
  (for 1--5 entries). Subagents carry the same spawning capability
  as their parent, enabling recursive decomposition bounded by a configurable
  depth limit.}
  \label{fig:rah_architecture}
\end{figure*}


\section{Introduction}
\label{sec:intro}

Modern coding agents operate within an \emph{agent harness}, the tools, filesystem, context
engineering, and orchestration that turn an LLM into a working agent
\cite{yang2024sweagent,jimenez2024swebench,yao2022react, lumer2026tool}. These harnesses navigate documents
far beyond a single context window
\cite{nakano2021webgpt,schick2023toolformer,qin2023toolllm}. Yet when a task requires
per-entry LLM reasoning across thousands of independent entries, the harness does not
\emph{recurse}. It cannot spawn subagents that carry its own tools
\cite{liu2023lostinmiddle,lee2024longcontext}. When an agent
recurses over a long-context task, should the recursive unit be a model call with no tools or a full
harness?

Each approach has a different blind spot. Coding agents reduce per-entry reasoning to regex heuristics.
Even with full filesystem access, their context windows cannot accommodate thousands of
entries simultaneously, forcing them to skip entry-level LLM reasoning entirely
\cite{sumers2023coala,mialon2023augmented}.
Conversely, recursive language models (RLMs) compose model calls recursively over sub-slices
of long inputs, performing reasoning that scales with the number of slices
\cite{zhang2024rlm}. RLMs name and validate recursion over model calls, and recent
production systems extend the same instinct to code that spawns subagents~\cite{anthropic2026workflows}.
We ask how far this pattern goes when the recursive unit is the full harness, and we name
and measure it. Benchmarks designed for long-context reasoning, such as Oolong,
expose the gap that motivates the comparison. Coding agents score low because of regex-limited
per-entry reasoning, while RLMs score low because of absent tool access.

We name this pattern the \textbf{Recursive Agent Harness (RAH)} and study it as a system
that spawns a full harness per entry through executable code
(Figure~\ref{fig:rah_architecture}). Consider a concrete instance, an Oolong document
containing 1{,}772 labeled key--value pairs spread across 536K tokens. A coding agent
writes a single script that loops regex matches over the document, and an RLM recursively
partitions the context but has no file access. RAH's parent agent instead writes a
script that issues one \texttt{Task()} per entry and runs them in parallel, and each call
resolves to an independent subagent harness with its own
context window, filesystem, and LLM call. Because the spawning logic is ordinary program code
rather than a fixed recursive-call convention or a schema-defined tool,
the parent can parametrize concurrency, per-entry instructions, and output paths in the same
language it uses for all other reasoning, and scale to entry counts that exhaust any
function-call budget.

RAH is evaluated on 199 randomly sampled instances from Oolong-Synthetic
\cite{cao2026oolong}, stratified across all 13 context-length buckets from 1K to 4M
tokens, using GPT-5 as the backbone to match the Codex baseline. With the model held fixed,
RAH improves the Codex result from 71.75\% to 81.36\%, which attributes the gain to the
harness rather than the model. Gains are consistent across all context-length buckets
including 4M tokens.

\paragraph{Scope of contribution.}
We make three contributions. First, we name \emph{Recursive Agent Harnesses} and define
\emph{harness recursion}, the use of a full agent harness rather than a model call with no tools as
the recursive unit, and situate it as the code-first extension to the model recursion of
RLMs~\cite{zhang2024rlm}. Second, we provide a controlled evaluation on long-context
reasoning with the backbone held fixed, so any difference is attributable to the harness
rather than the model. Third, we report results on Oolong-Synthetic across answer types and
context lengths, and relate the pattern to Anthropic's dynamic
workflows~\cite{anthropic2026workflows}, which adopt the same code-driven spawning in
production. The mechanism, code execution plus subagent spawning, is built from established
primitives~\cite{wang2024codeact,zhang2024rlm,wu2023autogen}. Our contribution is naming the
pattern and evaluating it, not inventing the primitives.


\section{Related Work}
\label{sec:related}

Cao et al.~\cite{cao2026oolong} apply a single coding agent that extracts per-entry answers with regex heuristics inside one context window, building off of previous work demonstrating the power of text-based search \cite{sen2026chronos, sen2026grep}. RAH instead spawns a subagent harness per entry or per group of entries, each with its own context window and tools, so a subagent can apply regex, reasoning, or both and cross-check the two rather than committing to regex alone.
Zhang et al.~\cite{zhang2024rlm} introduce recursive language models and establish that model recursion is an effective approach for long-context tasks, reaching 64.38\% on Oolong-Synthetic. RAH treats harness recursion as the natural complement to their model recursion, extending the same axis they open by adding tool access and per-entry subagent reasoning. RAH does not replace model recursion. It asks what changes when the recursive unit is upgraded from a model call with no tools to a full harness, an axis that RLMs opened.
Minions~\cite{narayan2025minions} and related orchestration frameworks fan work out to multiple worker models, and CodeAct~\cite{wang2024codeact} shows that writing and executing code is an effective agent action. RAH brings these together for long-context reasoning and provides the first controlled benchmark comparison against model recursion.
Anthropic's dynamic workflows let a coding agent write a script that orchestrates subagents at scale and execute the orchestration as code rather than turn by turn~\cite{anthropic2026workflows}. This is the same code-first spawning that RAH studies, and its arrival in a production system corroborates that recursion over agent harnesses is becoming a default strategy for tasks that exceed a single context window. We frame the pattern in the lineage of recursive language models rather than as a product workflow, treat it as model-agnostic and open to any harness, and contribute the controlled evaluation against model recursion that a product feature does not provide.

Lambda-RLM \cite{lambdarlm2026} decomposes long-context tasks through a deterministic pipeline with fixed structure and no tool access. RAH instead encodes spawning as an executable script that the parent generates at runtime, adapting structure to the workload rather than applying a predetermined schema.
AGENTHIVE \cite{agenthive2026} treats agent spawning as a first-class tool primitive defined in a schema. RAH encodes spawning directly in executable code, which lets the parent parametrize concurrency, output paths, and subagent instructions in the same language it uses for all other reasoning.
Parallel function calling \cite{kim2024llmcompiler} reduces latency by co-scheduling independent tool calls. RAH generalizes this idea to the level of entire agent invocations, spawning thousands of subagent harnesses rather than lightweight function calls.

Multi-agent conversation frameworks such as AutoGen \cite{wu2023autogen} and AgentVerse \cite{chen2023agentverse} coordinate agents through shared message threads. RAH instead isolates each subagent's context to prevent interference and enable deterministic aggregation through shared output files.
Embodied open-ended agents such as Voyager \cite{wang2023voyager} show that LLMs can write and execute programs as a form of persistent skill acquisition. RAH applies the same code-as-action insight to parallel document processing rather than sequential exploration.
MemGPT \cite{packer2023memgpt} addresses context limits by paging between main and external memory within a single agent. RAH instead distributes work across independent subagents, avoiding shared state entirely and scaling to entry counts that exceed what any paging scheme can address.
AgentBench \cite{liu2024agentbench} provides a multi-environment benchmark for evaluating LLMs as agents. Oolong complements AgentBench by focusing on aggregation over millions of tokens, a workload not covered by existing agent benchmarks.


\section{Recursive Agent Harness}
\label{sec:rah}

\subsection{Overview}
RAH makes the full agent harness, rather than a model call with no tools, the recursive unit (Figure~\ref{fig:rah_architecture}).
RAH treats code as a first-class action rather than a fixed tool schema, following the code-as-action view that writing and running a program is a more expressive way to orchestrate tools than emitting one structured call at a time~\cite{wang2024codeact}.
A parent agent receives the full task and inspects the document to determine workload size.
It spawns subagents through one primitive that the harness exposes two ways.
The baseline path is standard JSON tool calling, where the parent emits a structured call and the harness runs the subagent.
JSON tool calling is capped by the per-turn parallel tool-call budget.
RAH's contribution is the code-execution path, where the parent writes an executable script that instantiates subagents and runs them in parallel through its shell tool.
The code-execution path bypasses the per-turn cap and scales to thousands of subagent harnesses in parallel.
It is a harness primitive, not a system-prompt convention.
Every subagent carries the same tools and spawning capability as its parent, so the decomposition is recursive rather than one level of fan-out.

\subsection{Code-Execution Spawning}
\label{subsec:script}
For fine-grained workloads, the parent agent writes a self-contained script in which each subagent task is instantiated as a \texttt{Task()} object and all tasks are collected into a single asynchronous call that runs them in parallel. The number of subagents is set by the workload rather than any fixed cap, since it is simply a parameter in the generated code.
RAH is implemented with the LangChain agent framework package \cite{langchain2022}. The approach is otherwise independent of the implementation language. Our system emits Python and runs subagents under \texttt{asyncio.gather}, but any general-purpose language with concurrency primitives would serve.
The parent executes the script through its shell tool and receives only the aggregated stdout output once all subagents complete. Intermediate subagent reasoning, tool calls, and filesystem writes are invisible to the parent.
The script path sidesteps a practical constraint. JSON tool calling is bounded by the API's per-turn tool-call limit, whereas a script executed in a subprocess carries no such restriction, so spawning scale is controlled by the workload rather than the provider's protocol.
Each subagent is a full agent harness equipped with \texttt{read\_file}, \texttt{write\_file}, \texttt{ls}, \texttt{glob}, \texttt{grep}, \texttt{execute}, and web search, along with a planning step before execution. It operates inside an isolated workspace with no access to the parent context or to peer subagents.
The parent collects results by reading a shared output file that all subagents write to upon completion, avoiding any inter-process communication overhead.

\subsection{JSON Tool-Call Spawning}
\label{subsec:fc}
For subtasks of one to five entries, the parent calls the \texttt{Task} tool directly as a structured function call, passing entry content and instructions as arguments without generating an intermediate script.
Subagents spawned through JSON tool calling carry identical tool access and planning capability to those spawned through scripts, so the quality of per-entry reasoning is path-independent.
The parent selects the spawning path automatically based on entry count, ensuring that script-generation overhead is incurred only when the parallelism benefit justifies it.

\subsection{Subagent Architecture}
\label{subsec:subagent}
Subagents run an agent loop, alternating between planning, tool use, and reflection until a stopping condition is met \cite{wei2022cot,yao2023tot}.
Each subagent receives a bounded context with its assigned entry, relevant document excerpts, and output-format instructions. No shared memory or communication channel exists between sibling subagents, so each subagent reasons independently.
Each subagent is itself a full harness with the same tools, planning, and spawning capability as its parent, so the recursive unit is a harness rather than a model call. A subagent that meets a complex entry can write its own script and spawn grandchild harnesses, which makes the decomposition genuinely recursive rather than one level of fan-out.
Recursion depth is bounded by a configurable limit (default 3), which prevents unbounded tree growth while allowing multi-level decomposition when the workload warrants it.
The output of each subagent is a structured JSON record written to a designated path, which the parent script aggregates into a final answer after all subagent tasks resolve.


\section{Experiments}
\label{sec:experiments}

\subsection{Experimental Setup}

We evaluate harness recursion on long-context reasoning to answer two questions. First,
does making the harness the recursive unit improve over coding agents and over model
recursion? Second, how does performance vary by answer type and context length?
Oolong is a long-context reasoning benchmark that tests whether a model can aggregate
information distributed across millions of tokens rather than retrieve a single salient
span~\cite{bertsch2025oolong}, a setting on which frontier models score below 50\% at 128K
context. Evaluation uses 199 samples drawn from the Oolong-Synthetic validation split
\cite{cao2026oolong}, stratified across all 13 context-length buckets ranging
from 1K to 4M tokens (average 629K tokens per instance).
The sample size preserves the benchmark's bucket distribution while remaining
tractable under the per-instance agent cost.
RAH is compared against three baselines reported by Cao et al.~\cite{cao2026oolong}. These are a
full-context baseline that feeds the entire document to a frontier model
(Oolong Score 59.22\%), an RLM configuration (64.38\%), and a coding-agent
configuration using Codex-style reasoning with no retriever (71.75\%), which
represents the strongest prior result on this split. All prior results evaluate on the same 199-sample protocol.
Scoring follows the Oolong protocol, using exact match for LABEL, COMPARISON, USER,
and DATE answer types and $0.75^{|y - \hat{y}|}$ for NUMERIC responses, which
rewards near-miss predictions on continuous quantities. The reported Oolong Score is a
micro-average over the 199 instances. We pair point estimates with 95\% bootstrap
confidence intervals from 10{,}000 resamples of the per-instance scores.
RAH uses GPT-5 (\texttt{gpt-5-2025-08-07}) for the parent agent, every subagent, and the
answer-extraction call, matching the Codex baseline exactly, so any performance difference
is attributable to the harness architecture rather than model choice. The second
configuration uses Claude Sonnet 4.5 (\texttt{claude-sonnet-4-5-20250929}) as the backbone.
All runs use temperature zero. Final answers are extracted from subagent output through a
separate call using a fixed prompt asking for the answer in the exact format specified by
the question (Appendix~\ref{app:extract}). Because every answer passes through this GPT-5 extraction step while the
backbone under test is also GPT-5, the extractor and the system share a model family. Final
scoring is deterministic, using exact match and the numeric formula above rather than a model
judge, so the extraction step only maps raw output to the answer format and does not assign
scores. Raw outputs are typically already in that format, for example \texttt{Label: spam},
which limits the influence of the extractor. We did not run a separate human validation of the
extraction step and note it as a limitation.

\subsection{Main Results}

\begin{table}[t]
\centering
\caption{Oolong Score on Oolong-Synthetic (199 samples, 13 context-length
  buckets, GPT-5 backbone). Bold indicates best.
  The full-context, RLM~\protect\cite{zhang2024rlm}, and Codex baseline scores
  are all as measured and reported by Cao et al.~\protect\cite{cao2026oolong}.}
\label{tab:main_results}
\begin{tabular}{lc}
\toprule
Method & Oolong Score \\
\midrule
Full-context baseline & 59.22\% \\
RLM \cite{zhang2024rlm} & 64.38\% \\
Codex, No Retriever \cite{cao2026oolong} & 71.75\% \\
\midrule
RAH, GPT-5 (ours) & 81.36\% \\
\textbf{RAH, Sonnet~4.5 (ours)} & \textbf{89.77\%} \\
\bottomrule
\end{tabular}
\end{table}

\begin{figure}[t]
  \centering
  \includegraphics[width=\linewidth]{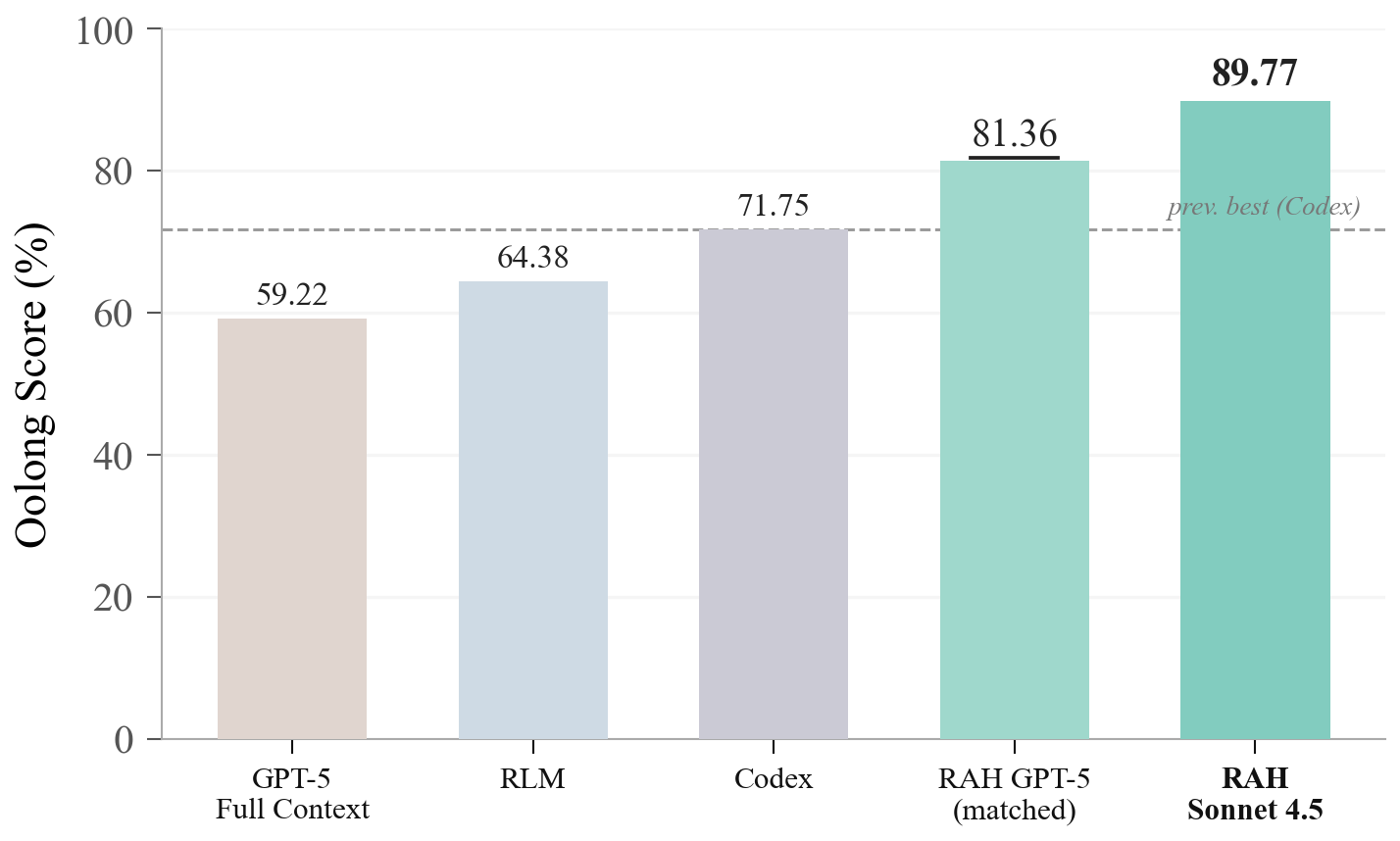}
  \caption{Oolong Score on Oolong-Synthetic (199 samples), mirroring
  Table~\ref{tab:main_results}. The three baselines all use a GPT-5 backbone and are
  reported by Cao et al.~\protect\cite{cao2026oolong}. With the backbone matched to Codex
  at GPT-5, RAH improves on the strongest prior result (dashed line), and the same design
  reaches 89.77\% on Claude Sonnet 4.5.}
  \label{fig:results}
\end{figure}

Table~\ref{tab:main_results} and Figure~\ref{fig:results} summarize the main results.
Holding the backbone fixed at GPT-5, RAH improves the strongest prior result,
the Codex coding agent, from 71.75\% to 81.36\%. Because both use GPT-5, this
controlled comparison attributes the difference to the harness rather than the
model. The same model that scores 71.75\% with a regex loop scores 81.36\% when
the recursive unit is a full harness. Treating the published baselines as fixed
reference points, since we do not have their per-instance scores, a bootstrap over the
199 RAH scores places the gain at $+9.61$ points over Codex (95\% CI $[4.2, 14.8]$) and
$+16.98$ over RLM (95\% CI $[11.5, 22.0]$), and both intervals exclude zero. The overall
score is 81.36\% (95\% CI $[76.0, 86.5]$). RAH also moves well past RLM, which reaches
64.38\% in this setting where it cannot open files, run code, or call external
services, showing that tool access and filesystem navigation complement recursive
LLM invocation. The design scales with the backbone as well, reaching 89.77\% on
Claude Sonnet 4.5, which suggests that harness recursion compounds with model
quality rather than substituting for it. Every RAH instance produced a
\texttt{Task()} script, so all samples followed the code-execution path, reflecting
that Oolong-Synthetic workloads uniformly exceed the five-entry threshold for the
JSON tool-calling path.

\subsection{Per-Category Analysis}

\begin{table}[t]
\centering
\caption{RAH Oolong Score by answer type on the 199-sample evaluation set.
  $n$ denotes the number of instances of each type. Intervals are 95\% bootstrap CIs.}
\label{tab:per_category}
\begin{tabular}{lccc}
\toprule
Answer Type & Score & 95\% CI & $n$ \\
\midrule
USER       & 87.27\% & [78.2, 94.5]  & 55 \\
COMPARISON & 89.29\% & [78.6, 100.0] & 28 \\
LABEL      & 86.54\% & [76.9, 94.2]  & 52 \\
DATE       & 60.00\% & [20.0, 100.0] &  5 \\
NUMERIC    & 69.33\% & [57.9, 80.1]  & 59 \\
\midrule
\textbf{Overall} & \textbf{81.36\%} & \textbf{[76.0, 86.5]} & \textbf{199} \\
\bottomrule
\end{tabular}
\end{table}

Table~\ref{tab:per_category} breaks results down by answer type.
Semantic types (USER, COMPARISON, LABEL) all exceed 86\%, confirming that
subagents produce reliable reasoning regardless of where in the
4M-token context the relevant key--value pairs reside.
NUMERIC performance degrades to 69.33\% because the $0.75^{|y - \hat{y}|}$
scoring function penalizes off-by-$k$ predictions compoundingly. An answer
off by one unit receives 0.75 and off by two receives 0.5625. This penalty
amplifies small counting errors into a visible score gap even when subagent
reasoning is correct. DATE ($n=5$) should be interpreted
cautiously given the small sample.

\subsection{Per-Context-Length Analysis}

\begin{table}[t]
\centering
\caption{RAH Oolong Score across all 13 context-length buckets ($n{=}14$--$16$ per bucket).
  The Codex baseline is 71.75\% flat across all buckets.}
\label{tab:per_length}
\begin{tabular}{lccc}
\toprule
Context & GPT-5 & Sonnet~4.5 \\
\midrule
1K   & 100.0\% & 93.8\%  \\
2K   &  93.8\% & 90.6\%  \\
4K   &  94.1\% & 94.1\%  \\
8K   &  94.5\% & 96.9\%  \\
16K  &  93.0\% & 90.6\%  \\
32K  &  86.7\% & 90.4\%  \\
64K  &  92.3\% & 100.0\% \\
131K &  73.3\% & 87.6\%  \\
262K &  57.1\% & 92.0\%  \\
524K &  80.0\% & 86.7\%  \\
1M   &  53.3\% & 80.0\%  \\
2M   &  66.7\% & 86.7\%  \\
4M   &  66.7\% & 76.7\%  \\
\bottomrule
\end{tabular}
\end{table}

Table~\ref{tab:per_length} reports scores across all 13 context-length buckets.
Sonnet~4.5 remains above 86\% through 524K tokens and above 76\% through 4M,
with no monotone degradation. The 64K bucket reaches 100\% and 262K outperforms
131K, which indicates that entry-count distribution within each bucket matters as much
as raw document length. GPT-5 follows the same broad pattern at lower magnitudes,
with a pronounced dip at 262K and 1M driven by NUMERIC-heavy instances. Both
configurations exceed the 71.75\% Codex baseline at the majority of context
lengths including 4M tokens. Per-bucket estimates rest on 14 to 16 instances and carry
wide intervals. The 1M bucket, for example, is 53.3\% with a 95\% CI of $[26.7, 80.0]$,
so we read bucket-level numbers as trends rather than precise points.

\subsection{Cost and Latency}
The cost of RAH scales with the number of subagents the parent chooses to spawn, which the
parent sets by deciding how many entries each subagent handles rather than fixing one
subagent per entry. The dominant recurring cost is each subagent re-reading the shared
document context. Prompt caching addresses this directly, since every subagent shares a
common context prefix, and prior work reports that caching can cut token cost by up to
80\% on long-horizon agentic workloads~\cite{lumer2026don}. Latency is bounded by
parallelism rather than by the count of subagents, because the parent's generated script
runs all subagents in parallel and waits only for the slowest
to return. We did not instrument exact token and wall-clock profiles for the GPT-5
configuration and leave precise cost characterization to future work.

\subsection{Failure Modes}
Three patterns account for most of the lost score. First, on a small number of instances
the parent answered directly without writing a spawning script, which collapses RAH to a
single coding agent and discards the per-entry reasoning the harness provides. These cases
concentrate at longer context lengths, where the parent is more likely to treat the task as
retrieval. Second, NUMERIC questions lose score even when subagent reasoning is sound,
because the $0.75^{|y-\hat{y}|}$ function penalizes an off-by-one count down to 0.75 and an
off-by-two count down to 0.5625. Third, answer types with few instances, such as DATE with
$n=5$, show high variance, so a single incorrect response moves the reported score by a
large amount and the 60.00\% DATE figure reflects sample size more than a systematic
weakness.

\subsection{Choosing a Recursion Strategy}
The recursive unit is the deciding factor when choosing among recursion strategies.
Table~\ref{tab:whentouse} contrasts the options. Coding agents pair a full harness with no
recursion and suit navigation over modest entry counts. Model recursion recurses over bare
model calls without tools and suits decomposition where tool use is unnecessary. Dynamic
workflows spawn subagents from a script and suit repeatable orchestration at scale. Harness
recursion makes the full harness the recursive unit, which suits per-entry reasoning at scale
where each entry may need tools.

\begin{table}[t]
\centering
\small
\caption{Recursion strategies by recursive unit.}
\label{tab:whentouse}
\begin{tabularx}{\linewidth}{@{} l >{\raggedright\arraybackslash}X >{\raggedright\arraybackslash}X @{}}
\toprule
\textbf{Strategy} & \textbf{Recursive unit} & \textbf{Strengths} \\
\midrule
Coding agent & none (single harness) & navigation, few entries \\
\midrule
Model recursion (RLM) & model call, no tools & decomposition without tools \\
\midrule
Dynamic workflows & subagent from a script & repeatable orchestration \\
\midrule
Harness recursion (RAH) & full agent harness & per-entry reasoning at scale \\
\bottomrule
\end{tabularx}
\end{table}

\section{Limitations}
\label{sec:limitations}

The evaluation is limited to Oolong-Synthetic. Generalization to Oolong-Real and to
domains where per-entry evidence is more ambiguous or less literally present in the
context remains an open question. The failure modes above also bound the result, in
particular the parent occasionally skipping spawning at long context lengths and the
NUMERIC scoring penalty that understates reasoning quality on continuous-quantity tasks.
RAH further depends on the parent agent reliably generating syntactically correct spawning
scripts, and exact token and latency profiles for the GPT-5 configuration remain
uninstrumented. We do not ablate individual design choices such as recursion depth, the
number of entries per subagent, or the code-execution versus tool-call spawning path.
Isolating their contributions is left to future work.


\section{Conclusion}
\label{sec:conclusion}

We name and evaluate the Recursive Agent Harness (RAH), the pattern of recursing over a full
agent harness rather than a model call with no tools. Long-context reasoning requires both
large-scale document navigation and per-entry LLM reasoning across thousands of independent
entries, and prior approaches provide only one. Coding agents reduce per-entry reasoning to
regex heuristics, while recursive language models lack filesystem access, code execution, and
external tools. Using GPT-5 to match the Codex baseline, RAH improves the strongest prior
result from 71.75\% to 81.36\% with no change in model, which attributes the gain to the
harness rather than the model. The pattern is already appearing in production systems such as
dynamic workflows. Our aim is to name it, place it in the lineage of recursive language models,
and measure it. The design should transfer to other settings that decompose into independent
per-entry subtasks over a corpus larger than the context window, such as multi-document
question answering and review of large document sets, which we leave to future work alongside
a characterization of the cost-quality tradeoff.




\section*{Code and Data Availability}
The Recursive Agent Harness implementation and the evaluation and scoring scripts will be
released shortly. The Oolong-Synthetic data and the published baseline numbers are due to
Cao et al.~\cite{cao2026oolong} and Zhang et al.~\cite{zhang2024rlm}. The parent, subagent,
and answer-extraction prompts are reproduced in full in Appendices~\ref{app:prompt},
\ref{app:subagent}, and \ref{app:extract}.

\bibliographystyle{ACM-Reference-Format}
\bibliography{references}

\appendix
\newpage
\section{Parent Agent Prompt}
\label{app:prompt}
The parent agent runs a general-purpose harness prompt. It contains no task- or
benchmark-specific instructions, and it does not tell the agent what to extract, how to score,
or how many subagents to spawn. Those decisions are left to the agent. We reproduce its
operative content below.

\begin{lstlisting}
You are a recursive agent harness. You have filesystem tools, code execution, and the
ability to spawn fresh sub-agent harnesses. Solve the task in the user message using
whatever combination of these is appropriate, and decide how to decompose the work
yourself.

Capabilities
- Filesystem: read_file, write_file, ls, glob, grep.
- Shell: run commands and scripts.
- Sub-agents: spawn a fresh agent harness for a self-contained subtask.

Spawning sub-agents
- For a few subtasks, call the Subagent tool directly.
- For many independent subtasks, write a short script that spawns sub-agents in parallel
  and run it with the Shell tool. Each Task(description=..., prompt=...) call runs one
  sub-agent and returns its output.

A minimal parallel example:

import asyncio
from recursive_agent_harness import Task

async def process(piece):
    result = await Task(description="subtask", prompt=f"...{piece}...")
    return result.output

async def main():
    pieces = [...]
    outputs = await asyncio.gather(*[process(p) for p in pieces])
    ...

asyncio.run(main())

Answer simple questions directly. Use sub-agents when the work decomposes into independent
parts.
\end{lstlisting}

\section{Subagent Prompts}
\label{app:subagent}
Spawned subagents are instantiated from one of three general-purpose harness types. The
parent selects the type, and the per-subtask instructions arrive in the user message. The system
prompts are reproduced below.

\begin{lstlisting}
general:
  "You are a focused sub-agent. Complete the task described in the user message.
   Be concise and return only the requested output. You have full filesystem access
   (read_file, write_file, ls, glob, grep) and can execute shell commands."

fast (read-only):
  "You are a fast read-only sub-agent. Answer the task in the user message directly
   and concisely. Use read_file, ls, glob, grep if you need to inspect data. Do NOT
   modify any files. Return only the requested output."

shell:
  "You are a command execution agent. Run the requested shell commands using the
   execute tool and report the output."
\end{lstlisting}

\section{Answer-Extraction Prompt}
\label{app:extract}
Each subagent's raw output is mapped to the answer format by a single follow-up call using
the prompt below, with a regular-expression parser as a fallback if that call returns empty.
On the 199-instance evaluation the extraction step succeeded on every instance, so the
fallback was never used. The extracted value is then compared to the ground-truth answer
programmatically, using exact match and the $0.75^{|y-\hat{y}|}$ numeric formula. No model
judge is used for scoring.

\begin{lstlisting}
The following is the output of an AI agent given this task:

TASK: {question}

AGENT OUTPUT:
{raw_output}

Extract ONLY the final answer in the exact format the task specifies.
Return just the answer value - no explanation, no punctuation, no extra words.
If the answer is a label like 'spam' or 'ham', return only that word.
If it is a number, return only the number. If it is a name or ID, return
only that value.
\end{lstlisting}


\end{document}